\documentclass[letterpaper, 10 pt, conference]{ieeeconf}  

\IEEEoverridecommandlockouts                              

\usepackage{caption}
\usepackage{latexsym}
\usepackage{booktabs}
\usepackage{graphicx}
\usepackage{color}
\usepackage{booktabs}
\usepackage{graphicx}
\usepackage{color}
\usepackage{subfigure}

\usepackage{appendix}
\usepackage{xcolor} 
\usepackage{ulem}
\usepackage{times}
\usepackage{latexsym}
\usepackage{amsmath}
\usepackage{multicol}
\usepackage{makecell}
\usepackage{graphicx,subfigure}
\usepackage{multirow}
\usepackage{balance}
\definecolor{mygreen}{RGB}{0,128,0}
\definecolor{mypurple}{HTML}{BF00BF}
\definecolor{mylawcase}{RGB}{127, 142, 186}
\definecolor{mylegal}{RGB}{241, 172, 106}
\usepackage{algorithm}
\usepackage{algorithmic}
\usepackage{multirow}
\usepackage{xcolor}
\usepackage{graphicx,subfigure}
\usepackage[export]{adjustbox}
\usepackage{microtype}
\usepackage{url}

\usepackage{cases}
\usepackage{color}
\usepackage[T1]{fontenc}

\usepackage{pslatex}
\title{\LARGE \bf
SEMDR: A Semantic-Aware Dual Encoder Model for Legal Judgment Prediction with Legal Clue Tracing
}
\author{Pengjie Liu$^{1}$, Wang Zhang$^{2}$, Yulong Ding$^{3}$, Xuefeng Zhang$^{4}$, Shuang-Hua Yang$^{5}$,~\IEEEmembership{Senior Member,~IEEE} 
\thanks{$^{1}$Pengjie Liu, $^{2}$Wang Zhang, $^{3}$Yulong Ding, and $^{5}$Shuang-Hua Yang are with School of Computer Science and Engineering, Southern University of Science and Technology, China.
        {\tt\small liupj2020@mail.sustech.edu.cn}; {\tt\small 12232429@mail.sustech.edu.cn}; 
        {\tt\small dingyl@sustech.edu.cn}; 
        {\tt\small yangsh@sustech.edu.cn}}%
\thanks{$^{4}$Xuefeng Zhang is with College of Science, Northeastern University, China.{\tt\small Zhangxuefeng@mail.neu.edu.cn}}%
\thanks{$^{5}$Shuang-Hua Yang is also with Department of Computer Science, University of Reading, UK.}%
}
\begin{document}

\maketitle
\thispagestyle{empty}
\pagestyle{empty}

\begin{abstract}
Legal Judgment Prediction (LJP) aims to form legal judgments based on the criminal fact description. However, researchers struggle to classify confusing criminal cases, such as robbery and theft, which requires LJP models to distinguish the nuances between similar crimes. Existing methods usually design handcrafted features to pick up necessary semantic legal clues to make more accurate legal judgment predictions. In this paper, we propose a Semantic-Aware Dual Encoder Model (SEMDR), which designs a novel legal clue tracing mechanism to conduct fine-grained semantic reasoning between criminal facts and instruments. Our legal clue tracing mechanism is built from three reasoning levels: 1) \textbf{Lexicon-Tracing}, which aims to extract criminal facts from criminal descriptions; 2) \textbf{Sentence Representation Learning}, which contrastively trains language models to better represent confusing criminal facts; 3) \textbf{Multi-Fact Reasoning}, which builds a reasons graph to propagate semantic clues among fact nodes to capture the subtle difference among criminal facts.
Our legal clue tracing mechanism helps SEMDR achieve state-of-the-art on the CAIL2018 dataset and shows its advance in few-shot scenarios. Our experiments show that SEMDR has a strong ability to learn more uniform and distinguished representations for criminal facts, which helps to make more accurate predictions on confusing criminal cases and reduces the model uncertainty during making judgments. All codes will be released via GitHub.
\end{abstract}
\section{Introduction}







Legal Judgment Prediction (LJP) is a semantic reasoning task in legal artificial intelligence. It aims to automatically and accurately predict the instrument labels (including: law imprisonment, legal charge, and law article)~\cite{CAIL2018}. The core requirement of LJP tasks is to recognize the necessary semantic clues from the criminal facts and regard them as supporting evidence to make legal judgments.
\begin{figure}[!t]        
    \centering
    \includegraphics[width = 0.9\linewidth] {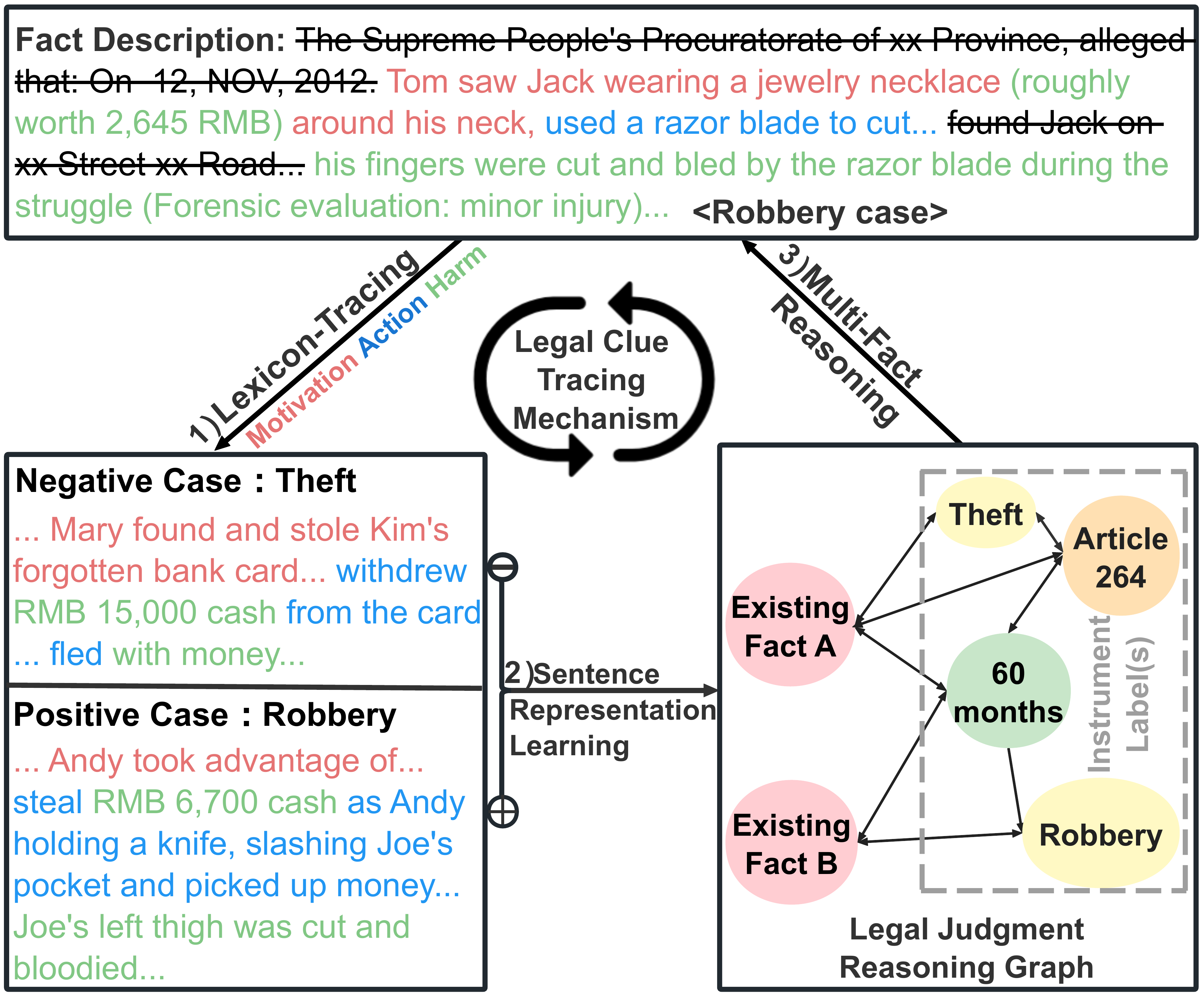}
    \caption{An Example of the Legal Clue Tracing Mechanism in SEMDR. We build a finer-grained reasoning framework for legal judgment prediction. The \sout{strikethrough part} in the fact description denotes the non-clue description, which will be filtered out during lexicon-tracing module.}
    \label{fig:1}
    \vspace{-4mm}
\end{figure}

Existing LJP models mainly focus on conducting fine-grained reasoning to make more accurate judgment predictions. Early work attempts to extract semantic facts of crimes by using template-based 
methods~\cite{kort1957predicting,ulmer1963quantitative,nagel1963applying} or legal handcrafted features and attributes~\cite{Liu2017,Yang2019,hu-etal-2018-shot} to distinguish different accusations. The development of pre-trained language models (PLMs), such as BERT~\cite{BERT-2019} and its variants, have yielded promising performance in conducting effective textual inference to make the judgments. To make PLMs better understand the semantics of legal fact, lots of work~\cite{zhong2019open,feng2022legal} continuously train language models by mask language modeling using legal domain corpus. Besides, some research also builds a reasoning graph over multiple criminal facts to conduct joint reasoning and enhance the representations of criminal cases~\cite{dong2021legal,yue-2021-neurjudge,li2021text}. Nevertheless, in the realistic legal judgment process, we usually conduct all these aspects to build a self-contained legal judgment prediction, including lexicon denoising, legal-semantic aware sentence representation, and multi-case reasoning.

In this paper, we propose a Semantic-Aware Dual Encoder Model (SEMDR), which designs a more effective clue tracing mechanism to conduct finer-grained reasoning in the LJP task. Our legal clue tracing contains three parts: Lexicon-Tracing, Sentence Representation Learning, and Multi-Fact Reasoning.
As shown in Figure \ref{fig:1}, we first pick up the criminal facts from the cases from motivation, action, and harm. Then SEMDR trains PLMs to learn more effective representations of the criminal fact to distinguish the subtle difference among confusing criminal cases. Precisely, we follow Gao et al.~\cite{gao2021simcse} and continuously train PLMs to represent criminal fact description by adding dropout noise. Finally, SEMDR builds a case enhancement graph to propagate the legal reasoning clues to the instrument label via the graph attention mechanism.
During contrastively training, our model adjusts its graph attention weights to dynamically pick up necessary semantic clues, which can support or refute the legal judgment predictions. Such an attention mechanism help to make the representations of instrument labels more distinguishable by aggregating these reasoning legal clues.



Our experiments show that SEMDR significantly outperforms existing LJP models with over 2.21\% improvements on confusing charge predictions from China Artificial Intelligence and Legal Challenges (CAIL2018)~\cite{CAIL2018}. Our further analyses illustrate that SEMDR conducts a self-contained clue tracing mechanism with the cooperation of the proposed modules of SEMDR. Our model showcases its advance on these low-frequency or confusing charges, which demonstrates its abilities of few-shot learning and fine-grained reasoning. Notably, our analyses explore that SEMDR can reduce the uncertainty during making legal judgment predictions, which thrives on the more distinguished and uniform embeddings of criminal cases and judgments. 
The main reason is that fine-grained legal clues can reduce the model uncertainty while making judgments by clustering the cases with corresponding judgment instruments and enabling the representations of confusing criminal cases to be more distinguished. 
\section{Related Work}
Earlier Legal Judgment Prediction (LJP) models usually aim to learn legal features from criminal cases for assisting legal judgment~\cite{haoxizhong-DBLP-2020}. However, the effectiveness of these models is limited by the quality of pre-defined trigger words and paradigms of legal judgment documents, resulting in that previous statistics-based LJP models constantly failing to distinguish low-frequency or confusing criminal charges~\cite{lauderdale2012supreme}.

Different from these feature-based LJP models, some researchers aim to use deep neural networks to extract legal elements from criminal facts in a fine-grained and automatic manner, leading to more accurate legal judgment predictions. Existing work~\cite{luo-etal-2017-learning, wang2018modeling, le2022legal, lyu2022improving} commonly utilizes the multi-head attention mechanism~\cite{vaswani2017attention} to capture semantics from legal cases and identify clues for making legal judgments.
However, these efforts primarily concentrate on specific LJP subtasks. Inspired by the dependencies observed in real-world, some research formulates the dependencies between subtasks to improve the model's performance~\cite{zhong-etal-2018-legal,ye2018interpretable,feng2022legal}.

Learning to effectively understand and represent legal text, such as criminal cases, law provisions, and legal charge definition, is crucial for LJP models to bridge the gap between textual description and their embedding in the semantic space.
For this reason, many works employ Pre-trained Language Models (PLMs) to encode legal texts for making judgments. 
They use mask language modeling to continuously train models to capture legal clues, understand legal semantics, and conduct more effective representations for these legal texts~\cite {ConSERT-2021,haoxizhong-DBLP-2020}. 
Nevertheless, the sentence embeddings encoded by PLMs are usually not sufficiently trained~\cite{li2020sentence,chen2021exploring,LEGAL-BERT-2020,ma2021legal,liu2024musemultiknowledgepassingedges}, making the performance of sentence representations show less effective in terms of modeling the semantic similarity of legal elements~\cite{Reimers-2019,xiao-Lawformer-2021}. Recent work constructs text pairs by adding several noise and contrastively trains PLMs to match these text pairs~\cite{fang2020cert,ConSERT-2021,CLEAR2020,liu2022augmenting,zhang2023contrastive,ge2021learning}. 
To represent legal texts and capture more fine-grained semantics to distinguish the difference between criminal cases, SimCSE~\cite{gao2021simcse} provides a promising way to learn sentence representations by directly adding dropout noise to sentence representations for contrastive training, making PLMs more sensitive to the subtle difference among similar sentence representations. Based on this, SAILER~\cite{li2023sailer} improves relevant legal case retrieval through a customized pre-training framework, ensuring more effective representation and retrieval of relevant/irrelevant cases.
Building an interpretable-perspective reasoning graph to propagate semantic information between evidence showcases remarkable efficacy in achieving more accurate text inference results~\cite{zhou2019gear,zhao2020transformer,liu2023ml}. They regard text segments as nodes, encode these textual descriptions with PLMs, and fully connect all nodes to build the text reasoning graph. Similarly, in the legal field, various graph modeling architectures can be employed to propagate semantic information between nodes in the customized graph~\cite{GCN,GraphSage,GAT}. Several researchers use defined legal keywords to build relations among criminal cases in a hierarchical graph for enhancing their representations~\cite{yue-2021-neurjudge,li2021text}. 
Additionally, to account for constraints and logical rules associated with some specific law articles, relational learning models are proposed to avoid unreliable judgments~\cite{dong2021legal,feng2022legal}. However, existing methods mainly focus on representing criminal cases while ignoring the learning of discriminative representations in instrument labels and their interaction, particularly for low-frequency and confusing criminal charges.
\section{Methodology}
In this section, we introduce the framework of Semantic-Aware Dual Encoder Model (SEMDR).
\begin{figure*}[t]   
    \centering
    \includegraphics[width =0.9\linewidth] {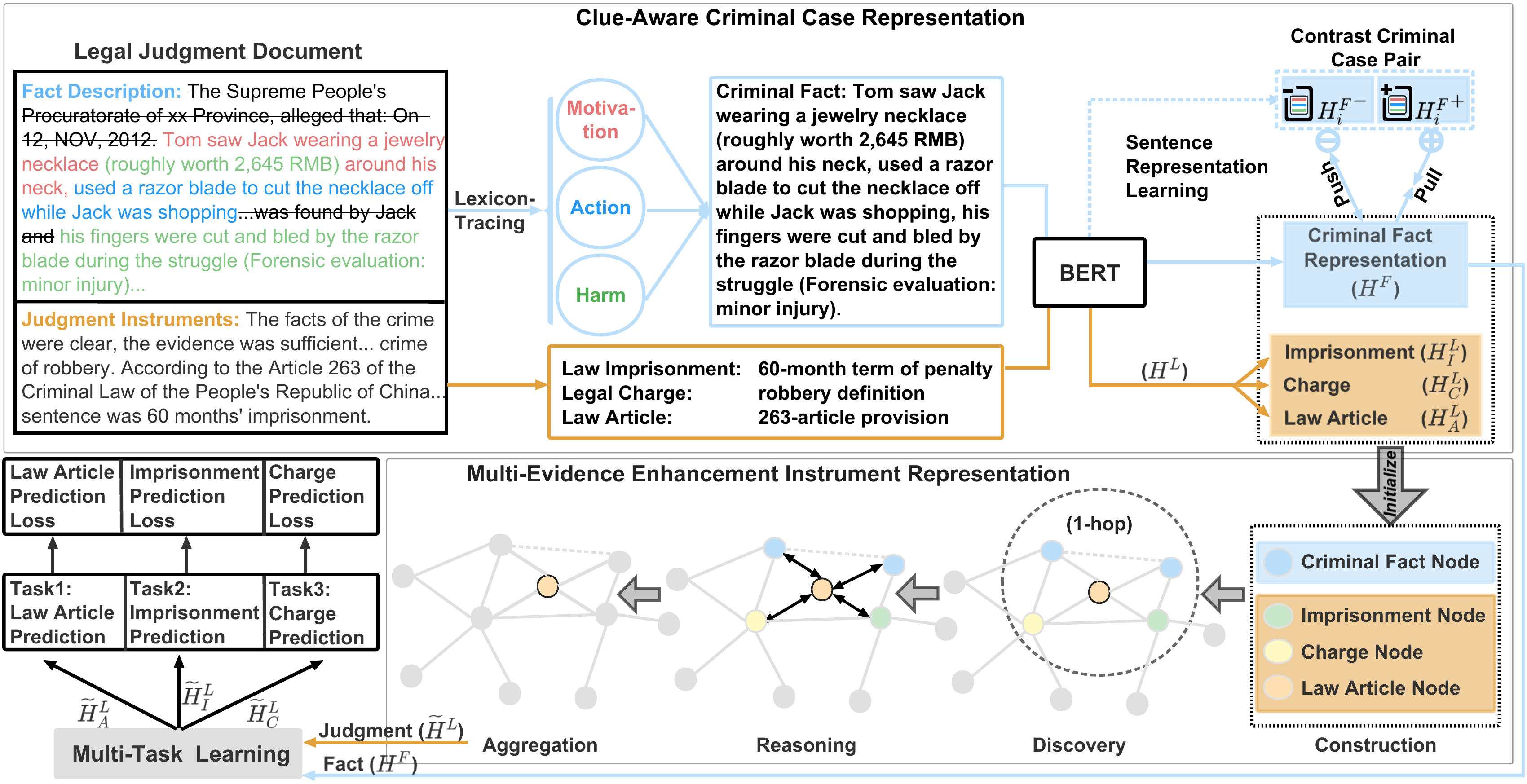}    
    \caption{ The Architecture of SEMDR Framework. In the legal judgment reasoning graph, we have defined several types of nodes ${(N)}$ and relations ${(R)}$ between criminal cases and corresponding judgment instruments.} 
    \label{fig:2}
    \vspace{-2mm}
\end{figure*}
\subsection{Preliminary of Legal Judgment Prediction}
Legal Judgment Prediction (LJP) aims to generate the unknown crime's verdicts according to the existing legal judgment documents. It includes three subtasks: \textbf{I}mprisonment prediction, \textbf{C}harge prediction, and Law \textbf{A}rticle prediction. We leverage criminal fact descriptions ($F$) to predict the corresponding judgment instrument labels ($I/C/A$).

As shown in Figure~\ref{fig:2}, SEMDR consists of two modules, including the clue-aware criminal case representation module, and the case enhancement graph reasoning module. 
We first conduct clue-aware criminal case representations~($H^{F}$) by employing pretrained language models (PLMs) to encode criminal cases and contrastively training these representations by adding dropout noise~(Sec. \ref{Sec.3.2}).
Then SEMDR establishes a legal judgment reasoning graph to enhance instrument label representation~($\widetilde{H}^{L}$) with the aggregated semantics from criminal cases~(Sec. \ref{Sec.3.3}).
Finally, SEMDR is trained to predict
the judgment label $L_{I/C/A}$ according to criminal fact representation $H^{F}$ using the following loss:
\begin{equation}
\small{\mathcal{L}_{\text{I/C/A}}{=}\mathrm{CrossEntropy}({L_{I/C/A}^\ast}, {{P({L_{I/C/A}|H^{F}})}})},
\end{equation}
where ${P({L_{I/C/A}|H^{F}})}$ denotes the probability to predict correct prediction results ${{L}}$, and ${L^\ast}$ represent the ground truth LJP verification labels. ${P({L_{I/C/A}|H^{F}})}$ is calculated by the similarity between the representation of 
criminal case~$H^{F}$ and instrument label~$\widetilde{H}_{I/C/A}^{L}$ as:
\begin{equation}
\small{P({L_{I/C/A}|H^{F}})=\mathrm{softmax}({sim(H^{F},\widetilde{H}_{I/C/A}^{L}))}},
\end{equation}
where the ${sim}$ denotes the dot product operation.
\subsection{Clue-Aware Criminal Case Representation}
\label{Sec.3.2}

Based on the research~\cite{haar_sawyer_cummings_1977,lauderdale2012supreme} in the legal field, effective and traceable legal clues can benefit the LJP models to identify confusing cases. 
Thus we extract motivation, action, and harm as the criminal fact rather than using the full text. Given the token sequence of criminal fact in the clue-aware criminal case representation module, we can obtain the corresponding representation of criminal fact~($H^{F}$):
\begin{equation}
\label{hf}
\small{H^{F}=\mathbf{BERT}\{{F^\mathrm{Motivation}, F^\mathrm{Action}, F^\mathrm{Harm}}\}.}
\end{equation}
Following previous work~\cite{gao2021simcse}, we add dropout noise to embedding representations and continuously train the criminal case representation ($H^{F}_{i}$) with its positive sample case ($H^{F}_{i^{+}}$) and negative sample case ($H^{F}_{i^{-}}$). The loss function of our contrastive learning~($\mathcal{L}_{\text{F}}$) is:
\begin{equation}
\label{CL}
\small{\mathcal{L}_{\text {F}}=-\log \frac{e^{\operatorname{sim}\left({H}^{F}, H^{F+}_{i}\right) / \tau}}{\sum_{i=1}^{N}\left(e^{\operatorname{sim}\left({H}^{F}, H^{F+}_{i}\right) / \tau}+e^{\operatorname{sim}\left({H}^{F}, H^{F-}_{i}\right) / \tau}\right)}.}
\end{equation}
We randomly sample cases as the negatives during contrastive learning. In addition, the ${sim}$ denotes the cosine similarity operation, and $\tau$ is the temperature hyperparameter.
\subsection{Multi-Evidence Enhancement Instrument Label Representation} 

\label{Sec.3.3}
To further enhance the representations of judgment instrument labels (${H}^{L}$), we build the legal judgment reasoning graph, 
where we define criminal cases and corresponding LJP labels as nodes and they are fully connected.
We apply the continuously trained PLM model in Sec.~\ref{Sec.3.2} to initialize the reasoning graph, and update the representation of the legal judgment node ($\widetilde{H}^{L}$) by relevant/similar cases aggregation.

\textit{Graph Building.}
To explicitly illustrate the association among criminal cases, we define criminal facts and legal instruments as nodes~(${N}$), and then design relations~(${R}$) to connect these nodes. In addition, the nodes of instrument labels are also connected internally.

\textit{Graph Initialization.}
We first encode the criminal fact $H^{F}$ in Eq.~(\ref{CL}) after contrastive learning, and initialize corresponding criminal fact nodes.

Then we split the judgment instruments of each criminal case into three independent parts: law imprisonment (${L_{I}}$), legal charge (${L_{C}}$) and law article (${L_{A}}$). Similarly, we leverage the BERT to encode the judgment instruments (${H^{L}}$) as:
\begin{equation}
    \small{H_{I/C/A}^{L} = \mathbf{BERT}(L_{{I/C/A}}).}
\end{equation}
These representations can initialize the legal judgement reasoning graph and then be enhanced by relevant cases.

Besides, SEMDR passes the information of connected nodes through cases aggregation, and the relational attention coefficient between nodes can be obtained through an activation function:
\begin{equation} 
\small{{R}_{{ij}}=\mathrm{LeakyReLU}\left({{\omega}}_{{ij}} \cdot [{{N}_{i}} \mid\mid {{N}_{j}}]\right),}
\end{equation}
 where $``\|"$ denotes the concatenation operation of nodes ${{N}_{i}}$ and ${{N}_{j}}$. We then use the dot product to calculate the learnable weight~(${\boldsymbol{{\omega}}_{{ij}}}$) and $\left[ {{N}_{i}} \|  {{N}_{j}}\right]$.
 
 \textit{Case Enhanced Reasoning.}
 Guided by the relations between criminal cases and their corresponding judgments, we leverage the Graph Attention Network (GAT)~\cite{vaswani2017attention} to conduct the graph representation and get the semantic-level attention ($\alpha_{i j}$):
\begin{equation}
    \small{\alpha_{{ij}}=\frac{\exp ({R}_{{ij}})}{\sum_{{e} \in \mathcal{O}({i})} \exp ({R}_{{ie}})},}
\end{equation}
where we take a softmax function to normalize the attention score. Moreover, $\mathcal{O}_{i}$ is the set of its 1-hop connected nodes.

Then the updated judgment results representation (${\widetilde{H}^{L}}$) can be enhanced by neighbours with $\alpha$: 
\begin{equation}
 \small{\widetilde{H}^{L}=\sigma\left(\sum_{j \in \mathcal{N}_{i}} \alpha_{i j} \cdot N_{i}\right),}
\end{equation}
where ${\sigma}$ is the ${ELU}$~\cite{clevert2015fast} activation function and 
neighbour nodes (${N_{i}}$) contain both criminal fact nodes and legal judgement nodes within discovery range.

\section{Experimental Methodology}
In this section, the experimental settings of SEMDR and baseline models are described.
\subsection{Datasets} 
We apply a large-scale public Chinese judgment dataset, CAIL2018, which contains the practice phase (CAIL-small) and the competition phase (CAIL-big). We also select low-frequency or occurrence highly confusing cases from both datasets to build the CAIL-mixed dataset. The statistics of datasets are listed in Table \ref{tab:1}.
\subsection{Baselines} 

We utilize several typical LJP models, including TF-IDF+SVM model, CNN-based models, LSTM-based models, and pre-trained language models based (PLM-based) models as baselines in our experiments:

\textit{TF-IDF+SVM model}: We apply the TF-IDF~\cite{SALTON1988513} to extract legal features, and combine the SVM~\cite{SVM} classifier to predict the corresponding judgments.

\textit{CNN-based models}: $\texttt{TextCNN}$~\cite{kim2014convolutional} has improved text classification performance with multiple filter widths. $\texttt{DPCNN}$~\cite{johnson2017deep} uses region embedding, convolution kernel and residual-connection to build the deep model. 

\textit{LSTM-based models}: 
$\texttt{LSTM}$~\cite{hochreiter1997long} builds a text classification model to predict verdicts.
\texttt{TopJudge}~\cite{zhong-etal-2018-legal} is a unified framework, which can deal with multiple LJP subtasks through topological learning on the Directed Acyclic Graph. \texttt{Few-Shot}~\cite{hu-etal-2018-shot} is the first multi-task model specifically designed to predict few-shot charges and confusing charges, combined with annotated legal attributes to enhance the keywords information in the criminal fact representation. \texttt{LADAN}~\cite{xu-etal-2020-distinguish} has devised a novel attentional mechanism that focuses on extracting salient features from fact description and automatically learns nuances between confusing legal text through a graph network.

\textit{PLM-based models}: \texttt{BERT}~\cite{BERT-2019} uses multiple encoders of Transformer and can capture bidirectional context information. \texttt{BERT-Crime}~\cite{zhong2019open} is fully pre-trained in large-scale corpus with legal data. \texttt{NeurJudge$^+$}~\cite{yue-2021-neurjudge} uses the predicted results of intermediate subtasks to divide factual descriptions into different situations and uses them to make predictions.
 \begin{table}[t]
\caption{The Statistics of CAIL2018 in Our Experiment.}
\resizebox{\linewidth}{!}{
\centering
\label{tab:1}
\centering
\begin{tabular}{lrrr} 
\hline
\textbf{Datasets}        & \multicolumn{1}{c}{\textbf{CAIL-small}} & \multicolumn{1}{c}{\textbf{CAIL-big}} & \multicolumn{1}{c}{\textbf{CAIL-mixed}}  \\ 
\hline
\text{Cases}           & 142,136                                  & 1,773,099                               & 102,661                                  \\
\text{Law Articles}    & 103                                      & 118                                   & 84                                      \\
\text{Charges}         & 119                                     & 130                                   & 82                                      \\
\text{Term of Penalty} & 11                                      & 11                                    & 11                                       \\
\hline
\end{tabular}
}
\vspace{-3mm}
\end{table}

\subsection{Testing Scenarios} 
Based on the frequencies of different types of legal judgment and miscarriage of justice shown in datasets, we define three criminal scenarios: high-frequency charge, low-frequency charge, and confusing charge scenarios.

We follow 
previous work~\cite{xu-etal-2020-distinguish} and keep the same settings\footnote{\url{https://github.com/prometheusXN/LADAN/tree/master}} in our experiments.
The high-frequency scenario contains criminal charges that have more than 100 cases while maintaining over 100 frequencies of law articles in CAIL-small and CAIL-big.
\begin{table*}[t]
\centering
\caption{Legal Judgment Prediction in the High-Frequency Criminal Charge Scenario (CAIL-small and CAIL-big). The best results are highlighted in \textbf{bold}, and the second-best results are \uline{underlined}.}
\label{cail-small}
\resizebox{\linewidth}{!}{
\begin{tabular}{l|ccc|ccc|ccc|ccc|ccc|ccc} 
\hline
Dataset                  & \multicolumn{9}{c|}{CAIL-small}                                                                                                                        & \multicolumn{9}{c}{CAIL-big}                                                                                                                                                                                                             \\ 
\hline
\multirow{2}{*}{Methods} & \multicolumn{3}{c|}{Law Article}                 & \multicolumn{3}{c|}{Charges}                     & \multicolumn{3}{c|}{Term of Penalty}             & \multicolumn{3}{c|}{Law Article}                                            & \multicolumn{3}{c|}{Charges}                                                & \multicolumn{3}{c}{Term of Penalty}                                          \\ 
\cline{2-19}
                         & Acc            & MP             & F1             & Acc            & MP             & F1             & Acc            & MP             & F1             & Acc                     & MP                      & F1                      & Acc                     & MP                      & F1                      & Acc                     & MP                      & F1                       \\ 
\hline
TF-IDF+SVM               & 81.56          & 78.68          & 73.89          & 80.20          & 79.04          & 75.16          & 37.61          & 36.71          & 31.93          & 93.17                   & 84.56                   & 77.10                   & 92.74                   & 82.64                   & 75.14                   & 49.24                   & 50.13                   & 50.14                    \\
LSTM                     & 86.69          & 79.99          & 78.43          & 85.95          & 81.81          & 81.07          & 40.35          & 39.61          & 33.96          & 95.41                   & 84.70                   & 84.61                   & 95.37                   & 85.41                   & 84.43                   & 54.46                   & 41.29                   & 38.32                    \\
DPCNN                    & 86.09          & 80.39          & 78.91          & 84.71          & 82.27          & 79.81          & 39.24          & 40.46          & 31.62          & 95.35                   & 86.57                   & 84.72                   & 95.21                   & 85.46                   & 84.47                   & 54.07                   & 41.17                   & 35.36                    \\
NeurJudge$^+$            & 84.17          & 80.99          & 79.86          & 87.40          & 83.28          & 81.71          & 37.64          & 36.43          & 35.33          & 94.44                   & 89.28                   & 87.04                   & 95.61                   & 92.12                   & 89.66                   & 53.82                   & 40.35                   & 38.17                    \\
TextCNN                  & 85.66          & 82.92          & 77.24          & 84.79          & 84.49          & 79.34          & 38.69          & 39.22          & 32.08          & 95.21                   & 88.91                   & 84.29                   & 95.06                   & 87.95                   & 83.55                   & 53.47                   & 42.49                   & 33.02                    \\
TopJudge                 & 87.04          & 85.41          & 80.59          & 85.84          & 85.67          & 81.16          & 38.59          & 36.80          & 31.49          & 94.28                   & 85.41                   & 80.59                   & 93.60                   & 85.67                   & 81.16                   & 52.77                   & 36.80                   & 34.68                    \\
LADAN                    & 89.24          & 84.95          & 78.79          & 86.89          & 82.90          & 82.84          & 40.19          & 34.69          & 35.62          & 95.48                   & 90.01                   & 90.08                   & 94.44                   & 87.81                   & 88.86                   & 53.26                   & 50.97                   & 51.77                    \\
Few-Shot                 & 86.32          & 85.97          & 78.69          & 85.95          & 84.35          & 81.57          & 37.56          & 34.66          & 36.33          & 89.63                   & 89.45                   & 89.41                   & 90.22                   & 90.35                   & 90.04                   & 50.01                   & 48.76                   & 49.42                    \\
BERT                     & 92.23          & 89.20          & 88.97          & 91.56          & 89.66          & 89.49          & 44.34          & 44.60          & 41.05          & 96.35                   & 95.82                   & 95.44                   & \uline{97.06}           & 96.37                   & 96.47                   & \uline{55.88}           & 54.96                   & 54.67                    \\
BERT-Crime               & \uline{92.54}  & \uline{89.64}  & \uline{89.15}  & \uline{91.88}  & \uline{90.01}  & \uline{90.44}  & \uline{44.59}  & \uline{44.78}  & \uline{41.14}  & \uline{96.37}           & \uline{96.07}           & \uline{95.58}           & 96.69                   & \uline{96.65}           & \uline{96.49}           & 55.59                   & \textbf{\textbf{55.20}} & \uline{54.94}            \\ 
\hline
\textbf{SEMDR}           & \textbf{95.48} & \textbf{95.25} & \textbf{94.07} & \textbf{94.74} & \textbf{93.59} & \textbf{94.86} & \textbf{46.35} & \textbf{44.94} & \textbf{43.98} & \textbf{\textbf{96.53}} & \textbf{\textbf{96.11}} & \textbf{\textbf{96.31}} & \textbf{\textbf{97.78}} & \textbf{\textbf{96.81}} & \textbf{\textbf{96.64}} & \textbf{\textbf{55.97}} & \uline{55.16}           & \textbf{\textbf{55.42}}  \\
\hline
\end{tabular}
}
\end{table*}
\begin{table}[!t]
\centering
\caption{Legal Charge Prediction in the Low-Frequency and Confusing Charge Scenarios. Specifically, the low-frequency charge scenario contains 72 types criminal charges in the CAIL2018 exercise phase, and three confusing charge scenarios all cover 10 high-frequency criminal charges.}
\resizebox{\linewidth}{!}{
\label{cail-big}
\begin{tabular}{l|cc|cc|cc|cc} 
\hline
Dataset                  & \multicolumn{8}{c}{CAIL-mixed}                                                                                                                     \\ 
\hline
\multirow{2}{*}{Methods} & \multicolumn{2}{c|}{Low-Frequency} & \multicolumn{2}{c|}{Confuse-Small} & \multicolumn{2}{c|}{Confuse-Medium} & \multicolumn{2}{c}{Confuse-Large}  \\ 
\cline{2-9}
                         & Acc            & F1                & Acc            & F1                & Acc            & F1                 & Acc            & F1                \\ 
\hline
TF-IDF+SVM               & 35.33          & 33.44             & 87.21          & 87.36             & 89.25          & 88.54              & 90.21          & 86.65             \\
LSTM                     & 43.55          & 44.20             & 63.91          & 62.37             & 75.62          & 73.25              & 93.91          & 93.88             \\
DPCNN                    & 44.36          & 44.45             & 87.20          & 87.27             & 89.69          & 88.21              & 93.56          & 93.54             \\
TextCNN                  & 45.69          & 48.36             & 88.91          & 88.93             & 90.22          & 87.23              & 93.97          & 93.79             \\
NeurJudge$^+$            & 52.19          & 52.04             & 86.81          & 87.37             & 88.63          & 85.39              & 92.13          & 90.68             \\
BERT                     & 52.26          & 54.23             & 92.19          & 92.23             & 93.22          & 92.55              & 94.48          & 94.47             \\
BERT-Crime               & 52.49          & 54.34             & \uline{92.37}  & \uline{92.41}     & \uline{93.33}  & \uline{92.69}      & \uline{95.03}  & \uline{94.91}     \\
TopJudge                 & 53.16          & 53.05             & 89.01          & 88.79             & 90.33          & 89.42              & 92.65          & 92.03             \\
LADAN                    & 55.08          & 55.30             & 88.89          & 88.72             & 90.48          & 89.51              & 93.05          & 91.33             \\
Few-Shot                 & \uline{55.23}  & \uline{55.98}     & 88.55          & 88.38             & 90.85          & 89.24              & 93.56          & 90.51             \\ 
\hline
\textbf{SEMDR}           & \textbf{57.33} & \textbf{57.89}    & \textbf{94.71} & \textbf{94.70}    & \textbf{95.55} & \textbf{94.61}     & \textbf{97.24} & \textbf{97.21}    \\
\hline
\end{tabular}
}
\end{table}
To verify that SEMDR has consistently excellent classification ability for some complex crimes, we collect charges with frequencies between 0 and 100 to form the low-frequency criminal charge scenario. Then we also select 10 criminal charges that could easily be misjudged to develop the confusing criminal scenarios (including: confuse-small, confuse-medium and confuse-large) where there are 50,000, 75,000, and 100,000 criminal cases, respectively.
\subsection{Implementation Details} 
There are two main modules in SEMDR, including clue-aware criminal sentence representation and graph reasoning.
To initialize clue-aware criminal sentence representation, we first apply Bert-base-Chinese to generate 256-dimension representation of criminal facts and three judgment instruments. In the legal judgment reasoning graph, we train 5000 epochs (high-frequency/confusing batch size: 20,000, low-frequency batch size: 1,200) for a 2-layer Graph Attention Network (GAT) to enhance instrument labels. Based on the best experimental results, we use the Adam optimizer with a learning rate of 0.01 and set the dropout probability to 0.5. 

In addition, for the CNN-based baselines, we set the maximum document length to 256 and used THULAC for word segmentation. Then for the LSTM-based models, we set the maximum number of words per sentence to 150 and the maximum number of sentences in a document to 15. For PLM-based models, we set the token limit to 512 per document during BERT embedding.

\textbf{Evaluation Metrics.} 
The official evaluation metrics of CAIL 2018\footnote{\url{ https://github.com/china-ai-law-challenge/cail2018 }} are used to evaluate LJP models, including Accuracy (Acc), Macro-Precision (MP), and Macro-F1 (F1). 


\section{Evaluation Results}
In this section, the experimental results are compared and analyzed to verify the effectiveness of SEMDR in Legal Judgment Prediction (LJP).
\subsection{Overall Performance}
We evaluate our model and baselines on all LJP subtasks and find SEMDR has achieved the best performance.

Table \ref{cail-small} and Table \ref{cail-big} exhibit SEMDR exceeds the baseline models in the high-frequency criminal charge scenario and improves more significantly on the CAIL-small dataset over baseline models. Specifically, on the CAIL-small dataset, SEMDR outperforms BERT-Crime on almost all subtasks by achieving over 5.61\% MP improvements in terms of law article prediction. Meanwhile, the PLM-based LJP models outperform the LSTM-based models, e.g., BERT-Crime achieves 10.12\% MP improvements than Few-Shot on imprisonment prediction. 
Analyzing the performance on the CAIL-big dataset, SEMDR continues to maintain a consistent improvement compared to the PLM-based models in all LJP subtasks, confirming the effectiveness of the legal clue tracing.
\subsection{Assessing the Performance of Legal Judgment Prediction Across Diverse Testing Scenarios}
We also have established two testing scenarios to evaluate the effectiveness of SEMDR, which are the low-frequency charge scenario and confusing charge scenarios.


As shown in Table \ref{cail-big}, SEMDR is benefited from the clue tracing mechanism and shows more accurate legal judgment predictions compared to other baselines in both low-frequency charge and confusing charge scenarios. It is worth noting that the LSTM-based models instead beat the PLM-based models, BERT-Crime, when making low-frequency charge prediction. On the contrary, in the confusing charge scenario, we find that the BERT-Crime performs best again among all baselines. The phenomenons show that BERT based LJP models are more effective in distinguishing the legal clues from confusing cases but still require more extensive data for finetuning. Furthermore, compared to BERT-based models, SEMDR has maintained consistent improvement and stable performance in all scenarios of different scales confuse-small, confuse-medium, and confuse-large. It demonstrates that SEMDR can effectively leverage a restricted amount of criminal supervision signals to produce convincing predictions.
\subsection{Ablation Study}
SEMDR consists of three main modules: lexicon-tracing, continuous training, and graph reasoning. In this subsection, we perform ablation studies to examine the individual contributions of each module toward the overall performance. 
SEMDR represents the criminal fact by leveraging two finer-grained legal text representation mechanisms, including lexicon-tracing and continuous training module. Then we use the graph reasoning module to enhance the legal judgment representation with the finer-grained case representations for clue tracing.
\begin{table}[!t]
\caption{Prediction Performance on Ablation Models.}
\resizebox{0.95\linewidth}{!}{
\centering
\label{tab4}
\begin{tabular}{l|cc|cc|cc} 
\hline
Dataset                                                                                              & \multicolumn{6}{c}{{CAIL-small}}                                                                                                                    \\ 
\hline
\multirow{2}{*}{Methods}                                                                             & \multicolumn{2}{c|}{{Law Article}}                  & \multicolumn{2}{c|}{{Charges}}                      & \multicolumn{2}{c}{{Imprisonment}}          \\ 
\cline{2-7}
                                                                                                     & {Acc}~                    & ~{F1}                     & {Acc}                     & {F1}                      & {Acc}~           & {F1}                       \\ 
\hline
{BERT (Dual-Tower Model)}                                                                                           & 60.22                   & 65.21                   & 63.72                   & 65.80~                  & 13.50          & 15.23                    \\
{+ Lexicon-Tracing}                                                                                     & 67.14                   & 70.39                   & 69.05                   & 70.10                   & 20.67          & 22.45                    \\
{+ Continuous Training}                                                                              & 83.61                   & 82.48                   & 80.59                   & 81.22                   & 38.75          & 38.55                    \\
{+ Graph Reasoning} & 88.01                   & 87.66                   & 87.68                   & 92.62                   & 40.66          & 38.45                    \\ 
\hline
{Full Model}                                                                                          & \textbf{\textbf{95.48}} & \textbf{\textbf{94.07}} & \textbf{\textbf{94.74}} & \textbf{\textbf{94.86}} & \textbf{43.35} & \textbf{\textbf{43.98}}  \\
\hline
\end{tabular}
}
\label{tab4}
\vspace{-2mm}
\end{table}

The evaluation results in Table \ref{tab4} show that all the modules can improve prediction performance. Further, we can observe that the graph reasoning module has achieved the most improvement (by ${27.79\%}$), which illustrates that the semantics from similar/related cases indeed helps to understand the LJP label to make more accurate judgments.

Then, our lexicon-tracing module grains a significant (5\%) accuracy improvement in each LJP subtask. This phenomenon illustrates that lexicon-tracing is a simple but effective method to denoise criminal cases to make more accurate prediction results.
Another significant improvement of SEMDR comes from the continuous training module. It helps PLMs to capture legal clues and conduct fine-grained criminal case representations to distinguish whether the representations are from different or the same criminal cases. 

The above three modules collaborate to achieve significant ($\textbf{35.26\%}$) accuracy improvements in the law article prediction, which shows the effectiveness of these modules in SEMDR.
\subsection{Visualization of SEMDR Behaviors}
In this subsection, we visualize the guidance of different SEMDR modules on the representations of criminal facts and instrument labels during confusing charge prediction.


\begin{figure}[!t]
  \centering
      \subfigure[BERT Embedding.]{\label{fig:4a}
            \includegraphics[width=0.18\textwidth]{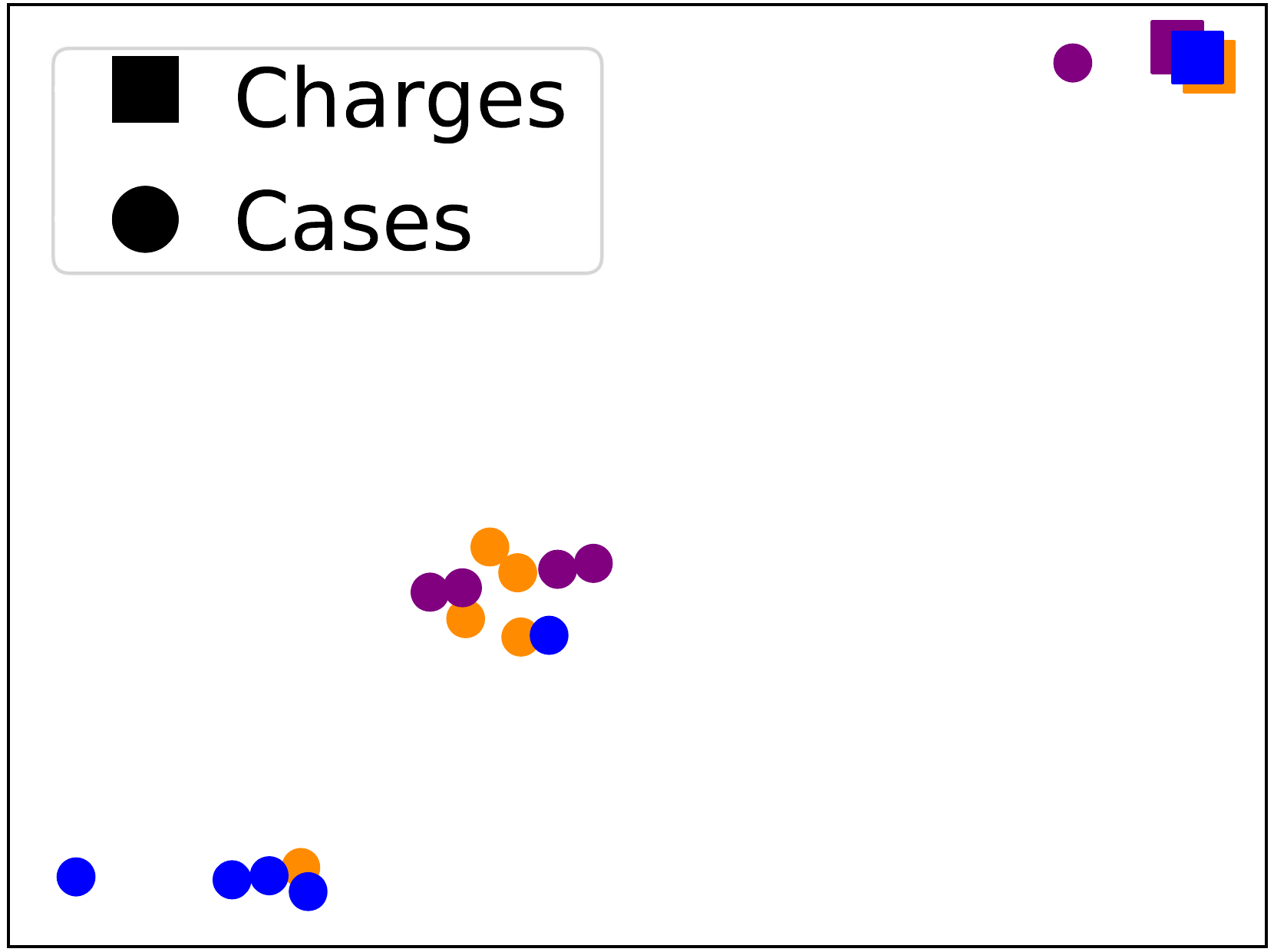}
            }
     \subfigure[SEMDR Embedding.]{\label{fig:4b}
            \includegraphics[width=0.18\textwidth]{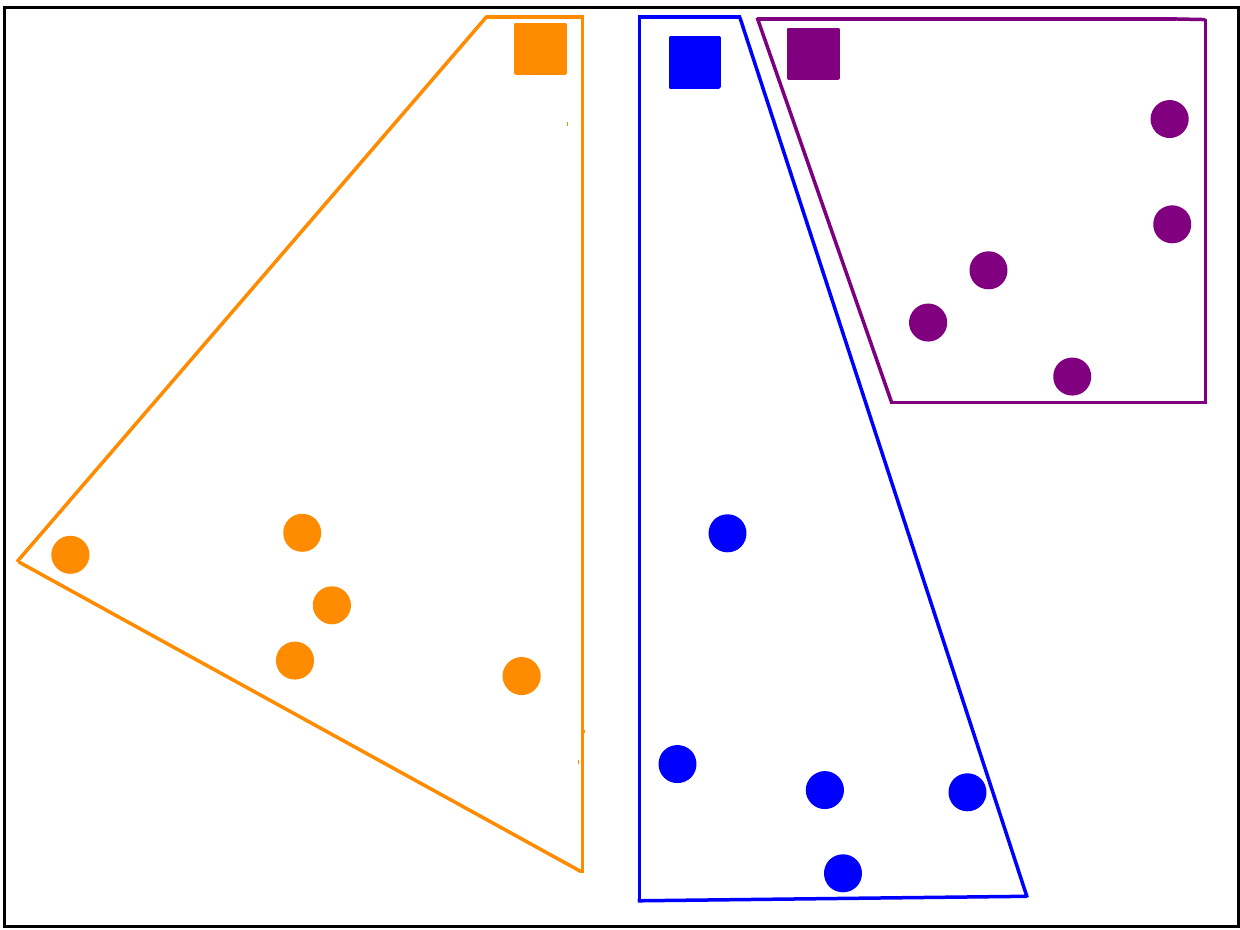}
            }
    \caption{Embedding Visualization (t-SNE) of Different Criminal Facts. The $``\;\rule[0.1em]{0.4em}{0.40em}\;"$ in different colours denotes the charges embedding of \textbf{<{\color{orange} robbery}, {\color{blue} theft}, {\color{purple}{fraud}}>} respectively and the ${``\bullet"}$ represents the fact embedding of their criminal cases. Figure \ref{fig:4a} to \ref{fig:4b} demonstrates the distribution of criminal cases and legal charges after different modules in SEMDR.}
\end{figure}
Then we choose three always confusing criminal charges (${<\textbf{robbery, theft, fraud}>}$), and randomly sample five corresponding criminal cases for each charge.
The t-SNE is employed to visualize the embedding distribution of their representations, and Figure \ref{fig:4a} shows a chaotic and collapsed embedding distribution encoded by the BERT. 
Specifically, the boundary between these confusing charges is ambiguous, and case nodes and charge nodes are mixed up, which fails to model the relevance between criminal cases and charges and distinguish the differences among criminal cases.
\begin{figure}[t]
    \centering
    \includegraphics[width=0.8\linewidth] {
    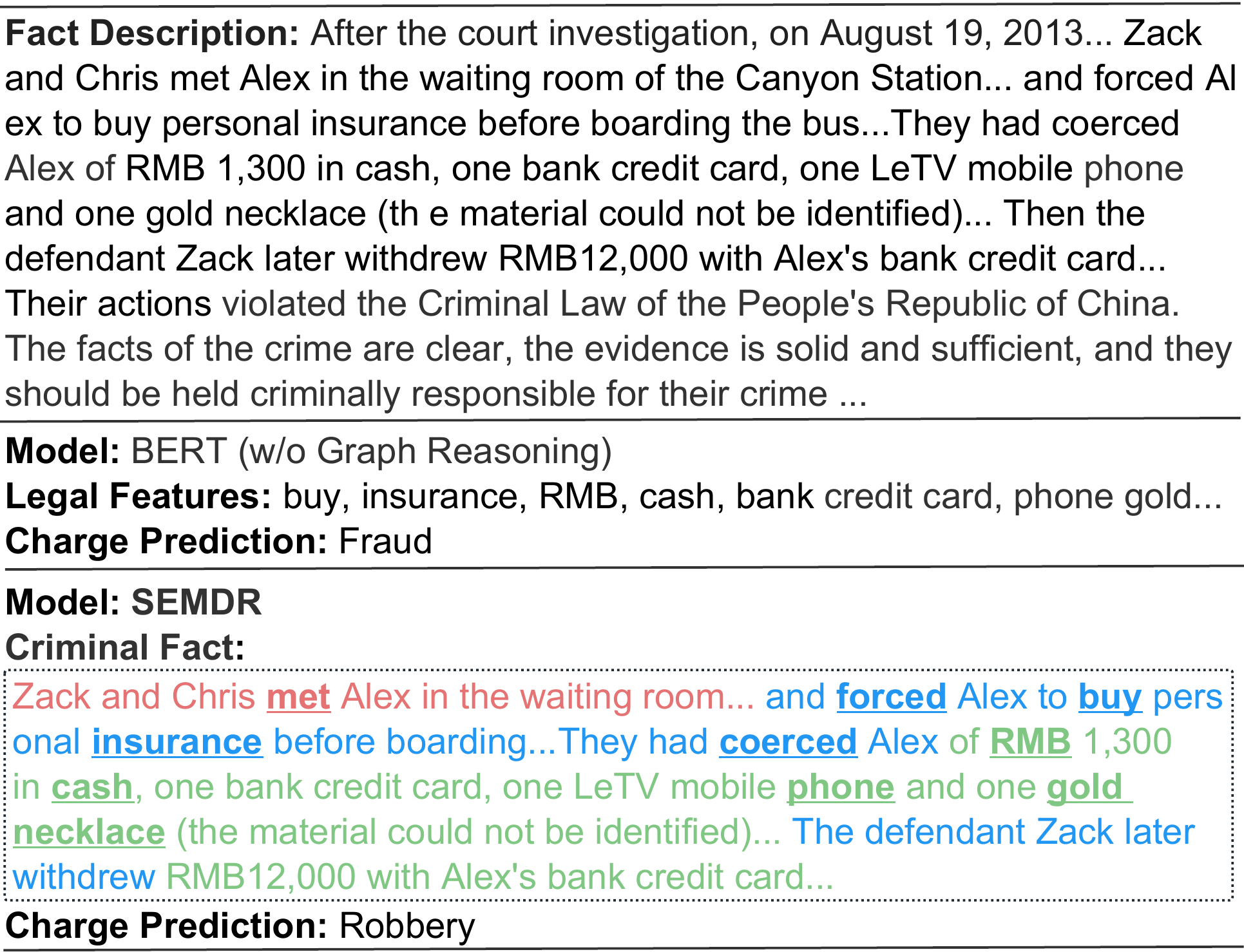
    }
    \caption{
    An Easily Misjudged Example Case Predicted by BERT (w/o Graph Reasoning) Model and SMEDR.
    }
    \label{fig:5}
    \vspace{-4mm}
\end{figure}
As shown in Figure \ref{fig:4b}, SEMDR calibrates embedding distributions and makes the embedding space more uniform and distinguished. Specifically, SEMDR clusters cases with their corresponding charges and makes the boundary clear of these confusing cases, which helps to make more accurate predictions.        

\subsection{Contributions of Legal Clue Tracing Mechanism}
This subsection demonstrates how the legal clue tracing mechanism works in SEMDR with a robbery case.

In Figure~\ref{fig:5}, we analyze a ``hard sell'' case involving a robbery charge. Although there are many legal features in the fact description, the BERT model without Graph Reasoning makes an erroneous prediction of fraud due to the ambiguity with other similar charges. 
However, SEMDR gives correct charge prediction after learning in both the clue-aware criminal case representation module and the multi-evidence enhancement instrument representation module.

\begin{figure}[t]{
\centering
\subfigure[
 The Comparison of Legal Charge Prediction Probability between BERT w/ Graph Reasoning and SEMDR.
 ]{
            \includegraphics[width = 0.8\linewidth]{
            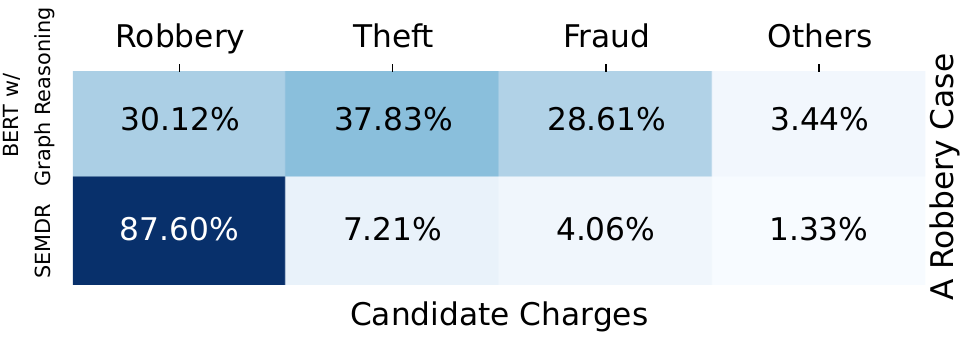
            }
            \label{fig:6a}
            }

  \subfigure[
  The Graph Attention Weights of Some Criminal Keywords and Charges after SEMDR Training.
  ]{
           \includegraphics[width = 0.8\linewidth] {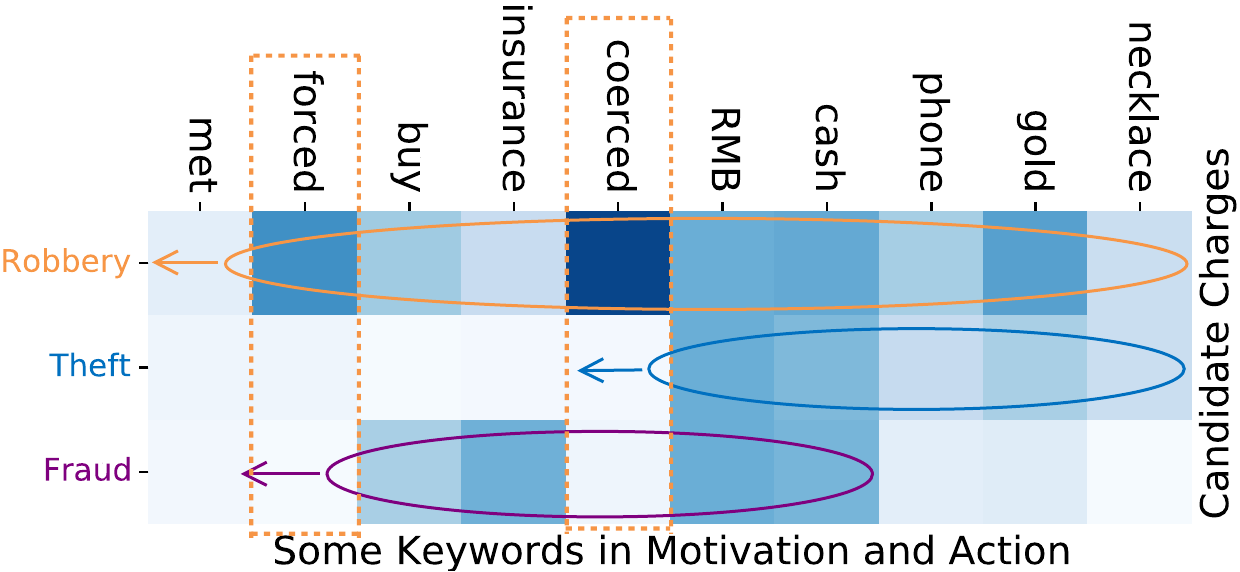}
            \label{fig:6b}}
    \caption{
    The Contributions of Legal Clue Tracing Mechanism to Assist SEMDR in Confusing Charge Prediction. A darker color means higher attention score and association.
    }
    }
    \vspace{-4mm}
\end{figure}

\textit{Contributions of Clue-Aware Criminal Case Representation.} 
As shown in Figure~\ref{fig:6a}, the BERT w/ Graph Reasoning model makes a wrong prediction as ``Theft'', which conducts a more uniform judgment prediction probability distribution on these confusing charge labels, theft (${37.83\%}$), robbery (${30.12\%}$), and fraud (${28.61\%}$). The main reason lies in that the BERT model captures more general legal clues ``money transactions'' to make a judgment prediction. On the contrary, SEMDR makes a more confident prediction of the proper charge, robbery ($\textbf{{87.60\%}}$), with the help of clue-aware criminal case representations.

\textit{Contributions of Multi-Evidence Enhancement Instrument Representation.} 
Then we further analyze the attention mechanism with the legal clue representations after the instrument label enhancement.
As shown in Figure \ref{fig:6b}, the crime of robbery receives more attention with the criminal fact than the other two charges. It proves that the representation of robbery learns more relevant criminal semantics during graph reasoning. 
In addition, we find that some keywords in motivation and action, such as forced and coerced, show significant and distinguished association with robbery, verifying that legal clue tracing can adjust graph attention weights, and dynamically select more core semantic signals.

\section{Conclusion}
In this paper, we propose a Semantic-Aware Dual Encoder Model, (SEMDR). SEMDR significantly reduces the uncertainty of the legal judgment prediction (LJP) model, and enhances the distinguish between legal charges guided by the legal clue tracing mechanism. Our experiments show that SEMDR achieves state-of-the-art performance, especially for the low-frequency/confusing criminal charges.


\begin{thebibliography}{99}
\bibitem{CAIL2018}Xiao, C., Zhong, H., Guo, Z., Tu, C., Liu, Z., Sun, M., Feng, Y., Han, X., Hu, Z., Wang, H. \& Xu, J. CAIL2018: A Large-Scale Legal Dataset for Judgment Prediction. { ArXiv Preprint}. (2018). 
\bibitem{kort1957predicting}Kort, F. Predicting Supreme Court decisions mathematically: A quantitative analysis of the “right to counsel” cases. { American Political Science Review}., 1-12 (1957).
\bibitem{ulmer1963quantitative}Ulmer, S. Quantitative analysis of judicial processes: Some practical and theoretical applications. { Law And Contemporary Problems}., (1963).

\bibitem{nagel1963applying}Nagel, S. Applying correlation analysis to case prediction. { Tex. L. Rev.}. pp. 1006 (1963)

\bibitem{Liu2017}Liu, Z. \& Chen, H. A predictive performance comparison of machine learning models for judicial cases. { IEEE Symposium Series On Computational Intelligence (SSCI)}. pp. 1-6 (2017)

\bibitem{Yang2019}Yang, W., Zhou, X. \& Luo, Y. Legal Judgment Prediction via Multi-Perspective Bi-Feedback Network. { Proceedings Of IJCAI}. (2019).
\bibitem{hu-etal-2018-shot}Hu, Z., Li, Liu, Z. \& Sun, M. Few-Shot Charge Prediction with Discriminative Legal Attributes. { Proceedings Of COLING}. (2018).

\bibitem{BERT-2019}Devlin, J., Chang, M., Lee, K. \& Toutanova, K. BERT: Pre-training of Deep Bidirectional Transformers for Language Understanding. { Proceedings Of NAACL-HLT}. pp. 4171-4186 (2019).

\bibitem{zhong2019open}Zhong, H., Zhang, Z., Liu, Z. \& Sun, M. Open chinese language pre-trained model zoo. { Technical Report}. (2019)

\bibitem{feng2022legal}Feng, Y., Li, C. \& Ng, V. Legal Judgment Prediction: A Survey of the State of the Art. { Proceedings Of IJCAI}. pp. 5461-5469 (2022)

\bibitem{dong2021legal}Dong, Q. \& Niu, S. Legal judgment prediction via relational learning. { Proceedings Of SIGIR}. pp. 983-992 (2021)

\bibitem{yue-2021-neurjudge}Yue, L., Liu, H., Zhang, K., An, Y., Cheng, M., Yin, B. \& Wu, D. NeurJudge: A Circumstance-aware Neural Framework for Legal Judgment Prediction. { Proceedings Of SIGIR}. pp. 973-982 (2021).

\bibitem{li2021text}Li, L., Bi, Z., Ye, H., Deng, S., Chen, H. \& Tou, H. Text-guided Legal Knowledge Graph Reasoning. { ArXiv Preprint}. (2021).

\bibitem{gao2021simcse}Gao, T., \& Chen, D. SimCSE: Simple Contrastive Learning of Sentence Embeddings. { Proceedings Of EMNLP}. pp. 6894-6910 (2021).

\bibitem{haoxizhong-DBLP-2020}Zhong, H., Xiao, C., Tu, C., Zhang, T., Liu, Z. \& Sun, M. How Does NLP Benefit Legal System: A Summary of Legal Artificial Intelligence. {Proceedings Of ACL}. pp. 5218-5230 (2020).

\bibitem{lauderdale2012supreme}Lauderdale, B. \& Clark, T. The Supreme Court's many median justices. { American Political Science Review}., 847-866 (2012).

\bibitem{luo-etal-2017-learning}Luo, B., Feng, Y., Xu, J., Zhang, X. \& Zhao, D. Learning to Predict Charges for Criminal Cases with Legal Basis. { Proceedings Of EMNLP}. pp. 2727-2736 (2017).

\bibitem{wang2018modeling}Wang, P., Yang, Z., Niu, S., Zhang, Y., Zhang, L. \& Niu, S. Modeling Dynamic Pairwise Attention for Crime Classification over Legal Articles. {Proceedings Of SIGIR}. pp. 485-494 (2018).
\bibitem{le2022legal}
Le, Y., Zhao, Y., Chen, M., Quan, Z., He, X. \& Li, K. Legal Charge Prediction via Bilinear Attention Network. \textit{Proceedings of the 31st ACM International Conference on Information \& Knowledge Management}. pp. 1024--1033 (2022).

\bibitem{lyu2022improving}
Lyu, Y., Wang, Z., Ren, Z., Li, H. \& Song, H. Improving legal judgment prediction through reinforced criminal element extraction. { Information Processing \& Management}. {59}, 102780 (2022).

\bibitem{vaswani2017attention}Vaswani, A., Shazeer, N., Parmar, N., Uszkoreit, J., Jones, L., Gomez, A., Kaiser, L. \& Polosukhin, I. Attention is All you Need. { Advances In Neural Information Processing Systems 30: Annual Conference On Neural Information Processing Systems 2017, December 4-9, 2017, Long Beach, CA, USA}. pp. 5998-6008 (2017).

\bibitem{zhong-etal-2018-legal}Zhong, H., Guo, Z., Tu, C., Xiao, C., Liu, Z. \& Sun, M. Legal Judgment Prediction via Topological Learning. { Proceedings Of EMNLP}. pp. 3540-3549 (2018).

\bibitem{ye2018interpretable}Ye, H., Jiang, X., Luo, Z. \& Chao, W. Interpretable Charge Predictions for Criminal Cases: Learning to Generate Court Views from Fact Descriptions. { Proceedings Of NAACL-HLT}.

\bibitem{ConSERT-2021}Yan, Y., Li, R., Wang, S., Zhang, F., Wu, W. \& Xu, W. ConSERT: A Contrastive Framework for Self-Supervised Sentence Representation Transfer. { Proceedings Of ACL}. pp. 5065-5075 (2021).

\bibitem{li2020sentence}Li, B., Yang, Y. \& Li, L. On the Sentence Embeddings from Pre-trained Language Models. { Proceedings Of EMNLP}, pp. 9119-9130 (2020).

\bibitem{chen2021exploring}Chen, X. \& He, K. Exploring simple siamese representation learning. { Proceedings Of CVPR}. pp. 15750-15758 (2021)

\bibitem{LEGAL-BERT-2020}Chalkidis, I., Fergadiotis, M., Malakasiotis, P., Aletras, N. \& Androutsopoulos, I. LEGAL-BERT: The Muppets straight out of Law School. { Findings of the EMNLP 2020}. pp. 2898-2904 (2020).

\bibitem{ma2021legal}Ma, L., Zhang, Y., Wang, T., Liu, X., Ye, W., Sun, C. \& Zhang, S. Legal judgment prediction with multi-stage case representation learning in the real court setting. { Proceedings Of SIGIR }. pp. 993-1002 (2021)

\bibitem{Reimers-2019}Reimers, N. \& Gurevych, I. Sentence-BERT: Sentence Embeddings using Siamese BERT-Networks. { Proceedings Of EMNLP}, (2019).

\bibitem{xiao-Lawformer-2021}Xiao, C., Hu, Tu, C. \& Sun, M. Lawformer: A pre-trained language model for chinese legal long documents. { AI Open}. pp. 79-84 (2021).

\bibitem{fang2020cert}Fang, H., Wang, S., Ding, J. \& Xie, P. Cert: Contrastive self-supervised learning for language understanding. { ArXiv Preprint}. (2020).

\bibitem{CLEAR2020}Wu, Z., Wang, S., Khabsa, M., Sun, F. \& Ma, H. CLEAR: Contrastive Learning for Sentence Representation. { ArXiv Preprint}. (2020).

\bibitem{liu2022augmenting}Liu, D., Du, W., Li, L., Pan, W. \& Ming, Z. Augmenting Legal Judgment Prediction with Contrastive Case Relations. { Proceedings of COLING}. pp. 2658-2667 (2022).

\bibitem{zhang2023contrastive}Zhang, H., Zhu, Y. \& Wen, J. Contrastive Learning for Legal Judgment Prediction. { ACM Transactions On Information Systems}. (2023).

\bibitem{ge2021learning}Ge, J., Huang, Y., Shen, X., Li, C. \& Hu, W. Learning fine-grained fact-article correspondence in legal cases. { IEEE/ACM Transactions On Audio, Speech, And Language Processing}. \textbf{29} pp. 3694-3706 (2021).

\bibitem{li2023sailer}Li, H., Ai, Q., Chen, J., Dong, Q., Wu, Y., Liu, Y., Chen, C. \& Tian, Q. SAILER: Structure-aware Pre-trained Language Model for Legal Case Retrieval. { Proceedings of SIGIR}, (2023).

\bibitem{zhou2019gear}Zhou, J., Han, X., Yang, C., Liu, Z., Wang, L., Li, C. \& Sun, M. GEAR: Graph-based Evidence Aggregating and Reasoning for Fact Verification. { Proceedings Of ACL}. pp. 892-901 (2019).

\bibitem{zhao2020transformer}Zhao, C., Xiong, C., Rosset, C., Song, X., Bennett, P. \& Tiwary, S. Transformer-XH: Multi-Evidence Reasoning with eXtra Hop Attention. { Proceedings Of ICLR}. (2020).

\bibitem{GraphSage}Hamilton, W., Ying, Z. \& Leskovec, J. Inductive Representation Learning on Large Graphs. { Advances in Neural Information Processing Systems}. pp. 1024-1034 (2017).

\bibitem{GAT}Velickovic, P., Cucurull, G., Casanova, A., Romero, A., Liò, P. \& Bengio, Y. Graph Attention Networks. { Proceedings Of ICLR}. (2018).

\bibitem{haar_sawyer_cummings_1977}Haar, C. \& Cummings, S. Computer Power and Legal Reasoning: A Case Study of Judicial Decision Prediction in Zoning Amendment Cases. { American Bar Foundation Research Journal}., 651-768 (1977).

\bibitem{clevert2015fast}Clevert, D., Unterthiner, T. \& Hochreiter, S. Fast and Accurate Deep Network Learning by Exponential Linear Units (ELUs). { Proceedings Of ICLR}. (2016).
\bibitem{SALTON1988513}
Salton, G. \& Buckley, C. Term-weighting approaches in automatic text retrieval. {Information Processing \& Management}, 513-523 (1988).

\bibitem{SVM}Suykens, J. \& Vandewalle, J. Least Squares Support Vector Machine Classifiers. { Neural Processing Letters}. pp. 293-300 (1999).

\bibitem{kim2014convolutional}Kim, Y. Convolutional Neural Networks for Sentence Classification. { Proceedings Of EMNLP}. pp. 1746-1751 (2014).

\bibitem{johnson2017deep}Johnson, R. \& Zhang, T. Deep Pyramid Convolutional Neural Networks for Text Categorization. { Proceedings Of ACL}. pp. 562-570 (2017).

\bibitem{hochreiter1997long}Hochreiter, S. \& Schmidhuber, J. Long short-term memory. { Neural Computation}., 1735-1780 (1997).

\bibitem{xu-etal-2020-distinguish}Xu, N., Wang, P., Chen, L., Pan, L., Wang, X. \& Zhao, J. Distinguish Confusing Law Articles for Legal Judgment Prediction. { Proceedings Of ACL}. pp. 3086-3095 (2020).

\bibitem{information-theory-2022}Pinkard, H. \& Waller, L. A visual introduction to information theory. { ArXiv Preprint}. (2022).

\bibitem{liu2024musemultiknowledgepassingedges}Liu, P. MUSE: Multi-Knowledge Passing on the Edges, Boosting Knowledge Graph Completion.  (2024), https://arxiv.org/abs/2408.05283

\bibitem{GCN}Kipf, T. \& Welling, M. Semi-Supervised Classification with Graph Convolutional Networks. {Proceedings Of ICLR}. (2017).

\bibitem{ZHANG2024110069}Zhang, X., Xiao, Z., Yang, B., Wu, X., Higashita, R. \& Liu, J. Regional context-based recalibration network for cataract recognition in AS-OCT. {Pattern Recognition}. {147} pp. 110069 (2024).

\bibitem{liu2023ml}Gan, Leilei and Kuang, Kun and Yang, Yi and Wu, Fei. Judgment prediction via injecting legal knowledge into neural networks. {Proceedings Of AAAI}. (2021).


\end{thebibliography}
\end{document}